\pdfoutput=1 

\documentclass[journal]{IEEEtran}
\ifCLASSINFOpdf
\else
\fi

\usepackage{amssymb}
\usepackage{mathtools}
\usepackage{amsmath}
\usepackage{algorithm}
\usepackage[noend]{algpseudocode}
\usepackage[pdftex]{graphicx} 
\usepackage{graphicx}
\usepackage{subfig}
\newcommand{\squeezeup}{\vspace{-2.5mm}}
\usepackage{color}

\usepackage[skip= 9pt, belowskip=-13pt,aboveskip=0pt]{caption}
\begin{document}
%
\title{Multi-agent Motion Planning for Dense and Dynamic Environments\\ via Deep Reinforcement Learning}
%
%
%

\author{Samaneh~Hosseini~Semnani, Hugh~Liu, Michael~Everett, Anton~de~Ruiter, Jonathan P. How
\thanks{“This work was supported by the Ontario Centers of Excellence, grant number 27481.” }
\thanks{S. Hosseini Semnani is with the Faculty of Department of Electrical and Computer Engineering, Isfahan University of Technology,Isfahan 15848-11888, Iran, and was with Aerospace Engineering Department, Ryerson University, Toronto, Canada (e-mail: samaneh.hoseini@cc.iut.ac.ir). }
\thanks{H. Liu is with the Faculty of Applied Science and Engineering, University of Toronto, Toronto, Canada (e-mail: liu@utias.utoronto.ca).}
\thanks{M. Everett is with the Aerospace Controls Laboratory, Massachusetts Institute of Technology, USA (e-mail: mfe@mit.edu). }
\thanks{A. de Ruiter is with the Aerospace Engineering Department, Ryerson University, Toronto, Canada (e-mail: aderuiter@ryerson.ca).}
\thanks{J. P. How is with the Aerospace Controls Laboratory, Massachusetts Institute of Technology, USA (e-mail: jhow@mit.edu). }
}

%
%

\markboth{}
{Shell \MakeLowercase{\textit{et al.}}: Bare Demo of IEEEtran.cls for IEEE Journals}
%



\maketitle

\begin{abstract}
This paper introduces a hybrid algorithm of deep reinforcement learning (RL) and Force-based motion planning (FMP) to solve distributed motion planning problem in dense and dynamic environments. Individually, RL and FMP algorithms each have their own limitations. FMP is not able to produce time-optimal paths and existing RL solutions are not able to produce collision-free paths in dense environments. Therefore, we first tried improving the performance of recent RL approaches by introducing a new reward function that not only eliminates the requirement of a pre supervised learning (SL) step but also decreases the chance of collision in crowded environments. That improved things, but there were still a lot of failure cases. So, we developed a hybrid approach to leverage the simpler FMP approach in stuck, simple and high-risk cases, and continue using RL for normal cases in which FMP can't produce optimal path. Also, we extend GA3C-CADRL algorithm to 3D environment. Simulation results show that the proposed algorithm outperforms both deep RL and FMP algorithms and produces up to 50$\%$ more successful scenarios than deep RL and up to 75$\%$ less extra time to reach goal than FMP.      
\end{abstract}


\IEEEpeerreviewmaketitle

\section{Introduction}
%
%
%
%
\IEEEPARstart{M}{ulti-agent} motion planning has recently attracted much interest in the research community and has many applications including robot navigation among pedestrians, self-driving cars, and drone shows. This problem concerns finding trajectories connecting each agent's initial location to its goal location. Each agent has a fixed final position that cannot be exchanged with another agent \cite{luis2019trajectory}. In addition, the motion planning algorithm should be able to satisfy constraints such as the maximum/preferred velocity of the agents and the minimum required separation distance between them. These constraints are dependent on the agents’ dynamic limitations and may vary from one application to another. In this problem agents make decisions in a dynamic environment with a (possibly varying) number of other agents, whose policies and intents are unknown.

Existing work on multi-agent motion planning problem can be broadly classified into two categories, centralized \cite{yu2016optimal, tang2018hold, augugliaro2012generation, preiss2017downwash} and decentralized approaches \cite{van2011reciprocala, van2011reciprocalb, alonso2013optimal, rezaee2013decentralized, SamaneFMP, zhou2017fast}. Centralized methods formulate the motion planning problem as an optimization problem in which information about position, velocity and goal location of all agents are available and the goal is to guide all agents toward their desired locations avoiding collision with each other and minimizing objectives such as energy or time. Yuin et al. in \cite{yu2016optimal} present a centralized algorithm based on linear programming (LP) to minimize last agent's arrival time, the maximum (single-agent) traveled distance, the total arrival time, and the total distance. Tang et al. in \cite{tang2018hold} decompose the problem into two phases: the motion planning step and the trajectory generation step. In the first step, a geometric algorithm finds piecewise linear trajectories for each robot. These trajectories are then refined in the second step into higher-order piecewise polynomials. Augugliaro et al. in \cite{augugliaro2012generation} cast the path planning problem into a non-convex optimization problem and use sequential convex programming to solve the problem. Centralized approaches suffer from issues in the scalability with the number of agents. The computational complexity of solving a large optimization problem is inevitable even if the problem is decomposed into several steps. In addition, for large-scale (many-agent) problems, it is often hard to establish a reliable communication network between all the agents and the central node. 

In contrast to centralized approaches, decentralized algorithms are properly scalable as they distribute the computational effort over multiple agents. They are also very robust to disturbance as they do real-time calculations. Optimal reciprocal collision avoidance (ORCA) \cite{van2011reciprocala} is one of the most commonly used algorithms in this category. In this algorithm, each agent, having the perfect knowledge about its neighbors' shape, position, and velocity, solves a local optimization problem to calculate a velocity that keeps it safe over the next time horizon. Original ORCA was able to work with holonomic robots only. Later, various variants of this algorithm proposed to work with the kinematics of non-holonomic robots \cite{snape2010smooth,alonso2013optimal}. Although ORCA guarantees collision-free navigation, its emergent behaviour is sensitive to hyper-parameter values.  Zhou et al. in \cite{zhou2017fast} introduced a method that lets each agent move just inside its calculated Voronoi cell and continues updating Voronoi cells until all agents reach their goal positions. This algorithm eliminates the requirement of neighbors' velocity information in ORCA. However, it still imposes high computational complexity of Voronoi cells calculation at each iteration over the agents and does not optimize the agents’ motions in terms of time, distance traveled or energy. Hosseini et al. in \cite{SamaneFMP} present FMP algorithm in which each agent, at each iteration, selects its next velocity by calculating the summation of some forces based on the flocking algorithm that guides the agent toward its goal location while avoiding collision with others. FMP is very fast and has low computational complexity overhead over the agents. Also, each agent in FMP only needs to know the relative position of its neighbors. However, like all decentralized algorithms, it will not produce a jointly time optimal trajectory because it is a decentralized algorithm in which none of the agents have a general view of the problem.
                
There are a number of recent studies looking to learning-based approaches for multi-agent motion planning problem. Learning-based techniques offload the expensive real-time motion planning computations to an off-line training procedure in which a policy function that implicitly encodes cooperative behavior among agents is learned \cite{muller2006off, everett2018motion, chen2017socially, long2018towards,pfeiffer2017perception,fujii1998multilayered}. The learned policy will be used by each agent later in real-time to select the best action at each time. Among learning based methods, RL approaches got the most interest as they do not require large training data sets, instead, they try to optimize the learned policy based on rewards they get by performing each action. Long et al. in \cite{long2018towards} present an end-to-end agent-free approach based on deep RL. They feed the raw Lidar sensor data to a trained network and receive each agents' steering command. However, in practice, it is useful to extract agent-level information from multiple sensors data to have more precise decisions. For example, if the Lidar sensor data show the presence of an object in the surrounding area, an agent reaction may differ if the object is another agent or if it is just an obstacle \cite{everett2018motion}. GA3C-CADRL \cite{everett2018motion} is another recent RL-based algorithm that has shown good performance in solving the path planning problem. However, RL approach presented in GA3C-CADRL is hard to train as its reward is sparse. Sparse reward in long-range navigation over complex maps is a challenge in many navigation problems using RL \cite{faust2018pearl, faust2018prm}. A bad reward function may lead to local minima or may lead an agent to wander around forever. To overcome this issue in long-range navigation tasks \cite{faust2018prm} presents a hybrid approach that combines sampling based path planning with RL. \cite{everett2018motion, chen2017socially} adds a separate pre-step supervised learning(SL) phase to the RL approach to teach agents how to reach their goal location without caring about colliding with other agents. Later they will learn how to avoid collision in RL phase. The main problem of this approach is its requirement of large training data sets which is not always easily available.

This research first presents a new goal-distance based proxy reward for the GA3C-CADRL algorithm to solve the multi-agent motion planning problem. This function not only eliminates the requirement of the pre-step SL phase but also reduces the number of collisions by introducing a more conservative strategy. The goal-distance based proxy reward had been shown to significantly improve RL performance for single agent problem aiming to find the shortest path toward the goal location \cite {ng1999policy}. The new reward function, in a multi-agent problem, not only attracts agents toward their goals but also prevents some collisions and converts some others to stuck and, as a result, increases the percentage of successful scenarios. Second, we present a hybrid control framework of combining deep RL and FMP algorithm called DRL-FMP in which, in normal scenarios, each agent selects its next action based on the learned policy by the presented extension of GA3C-CADRL (using the proposed new reward function) and switches to FMP algorithm in simple, stuck or high-risk scenarios. This work also extends GA3C-CADRL to solve 3D multi-agent planning problem. Simulation results are presented demonstrate the benefits of our hybrid control framework.  
\squeezeup
\section{Background}  
\subsection{Collision Avoidance with Deep RL (GA3C-CADRL)}
The multi-agent collision avoidance problem can be formulated as a sequential decision-making problem in an RL framework \cite{chen2017decentralized, chen2017socially, everett2018motion}. 
Let $\mathbf{s}_t$, $\mathbf{u}_t$, $\tilde{\mathbf{s}}_{j,t}$ denote an agent's state, action and the state of another agent ($j$) at time $t$. The state vector has two parts, the observable, $\mathbf{s}_{t}^{o}$ and unobservable, $\mathbf{s}_{t}^{h}$ parts, such that $\mathbf{s}_{t}=\left[\mathbf{s}_{t}^{o}, \mathbf{s}_{t}^{h}\right]$. The observable piece consists of the agent's position $\mathbf{p}$, velocity $\mathbf{v}$ and radius $r$, and is the part of the state that is visible by neighbouring agents. Therefore, each agent has local information about its neighbours. 
Other information, that is the goal position $\mathbf{p}_{g}$, preferred speed $v_{pref}$ and orientation $\psi$ for 2D environment and $\psi, \phi$ for 3D environment, are hidden from other agents and make up the unobservable part of the state. The action has two parts, speed $v_t$ and heading angle $\psi_t$ for 2D environment and $\psi_t, \phi_t$ for 3D environment. 
The objective for each agent $i$ is to minimize the expected time to reach goal $\mathbb{E}\left[t_{g}\right]$, while avoiding collision with its neighbouring agents ($N_i$), by developing a policy $\pi :\left(\mathbf{s}_{t}, \{\tilde{\mathbf{s}}_{j,t}^{o}\}\right) \mapsto \mathbf{u}_{t}   \:\:\:\:\:\: \forall j \in N_i$, satisfying
\begin{equation} \label{eq:1}
\underset{\pi\left(\mathbf{s}, \{\tilde{\mathbf{s}}_{j}^{o}\}\right)}{\operatorname{argmin}} \:\:\:\: \mathbb{E}\left[t_{g} | \mathbf{s}_{0}, \{\tilde{\mathbf{s}}_{j,0}^{o}\}, \pi\right] \:\:\:\:\:\:\: \forall j \in N_i \:\:\:\:\:\:\:\:\:\:\:\:\:\:\:\:\:\:\:\:\:\:\:\:\:
\end{equation}
\begin{equation} \label{eq:2}
s.t.  \left\|\mathbf{p}_{t}-\tilde{\mathbf{p}}_{j,t}\right\|_{2} \geq r+\tilde{r}_{j} \quad \forall t , \forall j \in N_i \:\:\:\:\:\:\:\:\:\:\:\:\:\:\:\:\:\:\:\:\:\:\:\:\:\:\:\:
\end{equation}
\begin{equation} \label{eq:3}
\mathbf{p}_{t_{g}}=\mathbf{p}_{g}\:\:\:\:\:\:\:\:\:\:\:\:\:\:\:\:\:\:\:\:\:\:\:\:\:\:\:\:\:\:\:\:\:\:\:\:\:\:\:\:\:\:\:\:\:\:\:\:\:\:\:\:\:\:\:\:\:\:\:\:\:\:\:\:\:\:\:\:\:\:\:\:\:\:\:\:\:\:\:\:\:\:\:\:\:\:
\end{equation}
\begin{equation} \label{eq:4}
 \mathbf{p}_{j,t}=\mathbf{p}_{j, t-1}+\Delta t \cdot \pi\left(\mathbf{s}_{j, t-1}, \{ \tilde{\mathbf{s}}_{k, t-1}^{o}\} \right)\:\: \forall t,  \:\: \forall j, \:\:\forall k \in N_j 
\end{equation}

where (\ref{eq:2}) is the collision avoidance constraint, $\Delta t$ is the time step, (\ref{eq:3}) is the goal constraint, (\ref{eq:4}) is the agents' kinematics, and the expectation in (\ref{eq:1}) is with respect to the other agent's unobservable states (intents) and policy. This simplified model of the agent is used to achieve faster computations. Higher-order models can be accommodated with minimum modification in what follows. An RL framework can be applied to generate a policy based on an agent's joint configuration with its neighbours, $\mathbf{s}^{j n}=\left[\mathbf{s}, \tilde{\mathbf{s}}^{o}\right]$. This RL framework applies a reward function, $R_{c o l}\left(\mathbf{s}^{j n}, \mathbf{u}\right)$, to penalize the agent in case of collision, and reward in case of reaching its goal. Two different types of RL algorithms are used in this RL framework, value-based \cite{chen2017decentralized, chen2017socially} and policy-based \cite{everett2018motion} learning. Value-based algorithm assumes that other agents continue their current velocities until next step, $\Delta t$, to be able to extract policy from the value function, $V\left(s_{t}^{j n}\right)$. This makes selecting the value of $\Delta t$ very critical in this approach. To avoid that, the GA3C-CADRL algorithm \cite{everett2018motion} applies GA3C \cite{babaeizadeh2016reinforcement}, a recent version of A3C algorithm \cite{mnih2016asynchronous}, to approximate both value and policy at the same time. A3C is an actor-critic algorithm with two, value and policy, losses
\squeezeup
\begin{equation} \label{eq:5}
Value\: Loss: f_{v}=\left(R_{t}-V\left(\mathbf{s}_{t}^{j n}\right)\right)^{2} \:\:\:\:\:\:\:\:\:\:\:\:\:\:\:\:\:\:\:\:\:\:\:\:\:
\end{equation}
\begin{equation} \label{eq:6}
\begin{aligned}
Policy\: Loss: f_{\pi}=\log & \pi\left(\mathbf{u}_{t} | \mathbf{s}_{t}^{j n}\right)\left(R_{t}-V\left(\mathbf{s}_{t}^{j n}\right)\right)\\
      & + \beta \cdot H\left(\pi\left(\mathbf{s}_{t}^{j n}\right)\right)
\end{aligned}
\end{equation} 
 
where $\gamma$ is discount factor, $R_{t}=\sum_{i=0}^{k-1} \gamma^{i} r_{t+i}+\gamma^{k} V\left(\mathbf{s}_{t+k}^{j n}\right)$ is a discounted reward estimate, $\beta$ is a tunable constant parameter and $H$ is the entropy function. Equation (\ref{eq:5}) is used for training the value function to match with $R_{t}$. Equation (\ref{eq:6}), in its first term, penalizes actions with high probability that leads to lower rewarded estimate than what was predicted by value function. The second term uses the entropy as a means of encouraging exploration. If the policy outputs actions with relatively similar probabilities, then the entropy will be high, but if the policy suggests a single action with a large probability then the entropy will be low. In A3C multiple threads of an agent interact with their own copy of the environment at the same time as the other threads are interacting with their environments. At the end of each round, each thread updates global network parameters using the calculated gradients by its loss terms. GA3C is an extension of A3C that uses queues for training and prediction purposes, to efficiently use the GPU and as a result, learns faster. Our work builds on open-source cadrl-ros package \cite{cadrlros} that implements GA3C-CADRL algorithm \cite{everett2018motion}.  
\squeezeup        
\subsection{Force-based Motion Planning (FMP)}
The Force-based motion planning algorithm \cite{SamaneFMP} is an extension of the Flocking algorithm \cite{semnani2014semi,olfati2004flocking} that is designed to work as a motion planning algorithm. To reach this point, FMP changes the three main forces of the Flocking algorithm: gradient-based, velocity consensus and navigational feedback. The gradient-based term is changed such that it has only repulsive force instead of the original attractive/repulsive force of the Flocking algorithm. The Velocity consensus term is omitted because FMP does not need to keep the agent's velocity close to their neighbors. Finally, the navigational term of FMP changed to attract each agent to its own specific target position.
In FMP, each agent repeats calculating and applying a control input function  
\squeezeup
\begin{equation} \label{eq:7}
u_{i}=\left\{\begin{array}{ll}{0} & {\text { if } v_{i}^{T} \overline{u}_{i}>0 \text { and }\left\|v_{i}\right\| \geq V_{\max }} \\ {\overline{u}_{i}=f_{i}^{\mathcal{R}}+f_{i}^{\gamma}} & {\text { otherwise }}\end{array}\right.
\end{equation}      

where $f_{i}^{\mathcal{R}}$ and $f_{i}^{\gamma}$ are the repulsive and navigational forces respectively. $v_{i}$ is velocity of agent $i$ and $V_{\max }$ is the maximum allowed velocity of that agent.  Equation (\ref{eq:7}) is applied, until all the agents reach their goal locations or the algorithm time outs. FMP has a version for obstacle avoidance and the mathematical analysis presented in \cite{SamaneFMP} shows if the initial position of the agent is large enough then the algorithm is always able to keep a required distance between the agents and the algorithm is proved to be live-lock free. Although FMP is real-time, distributed, fast and scalable, it is not able to provide a time optimal path. Therefore, in this research, we focus on a modified version of GA3C-CADRL as the basic algorithm to provide closer to optimal paths and switch to FMP in case of stuck scenarios and some other situations that are described in Section \ref{sec:3-3}.     
 \squeezeup
 \squeezeup
\section{Approach}   
\subsection{GA3C-CADRL-NSL}
GA3C-CADRL outperforms the previous versions of the CADRL algorithm as the number of agents in the environment grows, however it still suffers from two main drawbacks. First, it requires a pre-step SL training. Second, its performance in dense problems drops greatly. This section presents an extension of GA3C-CADRL, called GA3C-CADRL-NSL (GA3C-CADRL-No Supervised Learning), that uses a new reward function. This new function first eliminates the requirement of pre-step SL training. Second as it is more conservative,  the number of successful scenarios increase. Pre-step SL requires gathering a large set of training data which is not always easy.
The reward function used in GA3C-CADRL is given by
\squeezeup
\begin{equation} \label{eq:8}
R_{c o l}\left(\mathbf{s}^{j n}\right)=\left\{\begin{array}{ll}{1} & {\text { if } \mathbf{p}=\mathbf{p}_{g}} \\ {-0.25} & {\text { if } d_{\min }<0} \\ {-0.1+0.05 \cdot d_{\text { min }}} & {\text { if } 0<d_{\min }<0.2} \\ {0} & {\text { otherwise }}\end{array}\right.
\end{equation}
where $d_{min}$ is the distance to the closest other agent. This reward function has two parts: goal-reward and collision-punishment. Goal-reward only gives reward 1 to the agent when it reaches its goal location, before then, the agent won't get any positive reward. This is the reason that GA3C-CADRL requires an initial SL step. Because non-initialized agents wander randomly and probably won't ever reach their goal to get any posetive reward. Therefore, GA3C-CADRL applies SL to teach agents on how to reach their goals, then applies RL to teach them how to avoid collision. GA3C-CADRL-NSL changes the goal reward to attract agents toward their goal location inside RL which eliminates the requirement of SL. GA3C-CADRL-NSL's reward function is given by
\squeezeup
\begin{equation} \label{eq:9}
R\left(s^{j n}\right)=\text {R}_{\text {c}}+\text {R}_{\text {g}}
\end{equation}
\squeezeup
\begin{equation}\label{eq:10}
\text {R}_{\text {c}}=\left\{\begin{array}{ll}{-1} & {\text { if } d_{\min }<0} \\ {10 . d_{\min }-1} & {\text { if } 0<d_{\min }<0.1} \\ {0} & {\text { otherwise }}\end{array}\right.
\end{equation}
\squeezeup
\begin{equation}\label{eq:11}
\text {R}_{\text {g}}=\left\{\begin{array}{ll}{1} & {\text { if } p=p_{g}} \\ {\propto\left(\text {goal}_{\text {dist}}^{t-1}-\text {goal}_{\text {dist}}^{t}\right)} & {\text { otherwise }}\end{array}\right.
\end{equation} 

where $\text {goal}_{\text {dist}}^{t}$ is the Euclidean distance between agent and its goal at time $t$ , $\mathrm{R}_{\mathrm{c}}$ is  collision punishment and $\text {R}_{\text {g}}$ is the goal reward. $\propto$ is a constant that is set to $0.08$ in the both 2D and 3D training procedures. $\text {R}_{\text {g}}$ gives a small positive/negative reward to the agent by coming closer/further to its goal respectively. This prevents agents from wandering randomly and guide them toward their goal.
The new reward function also provides more conservative behavior by changing collision punishment to give the maximum penalty, -1,  in case of collision and linearly decreasing that punishment until $d_{min}=0.1$.   
\squeezeup
\subsection{Training the Policy}
\begin{figure}[!t]
\centering
\includegraphics[width=2.5in]{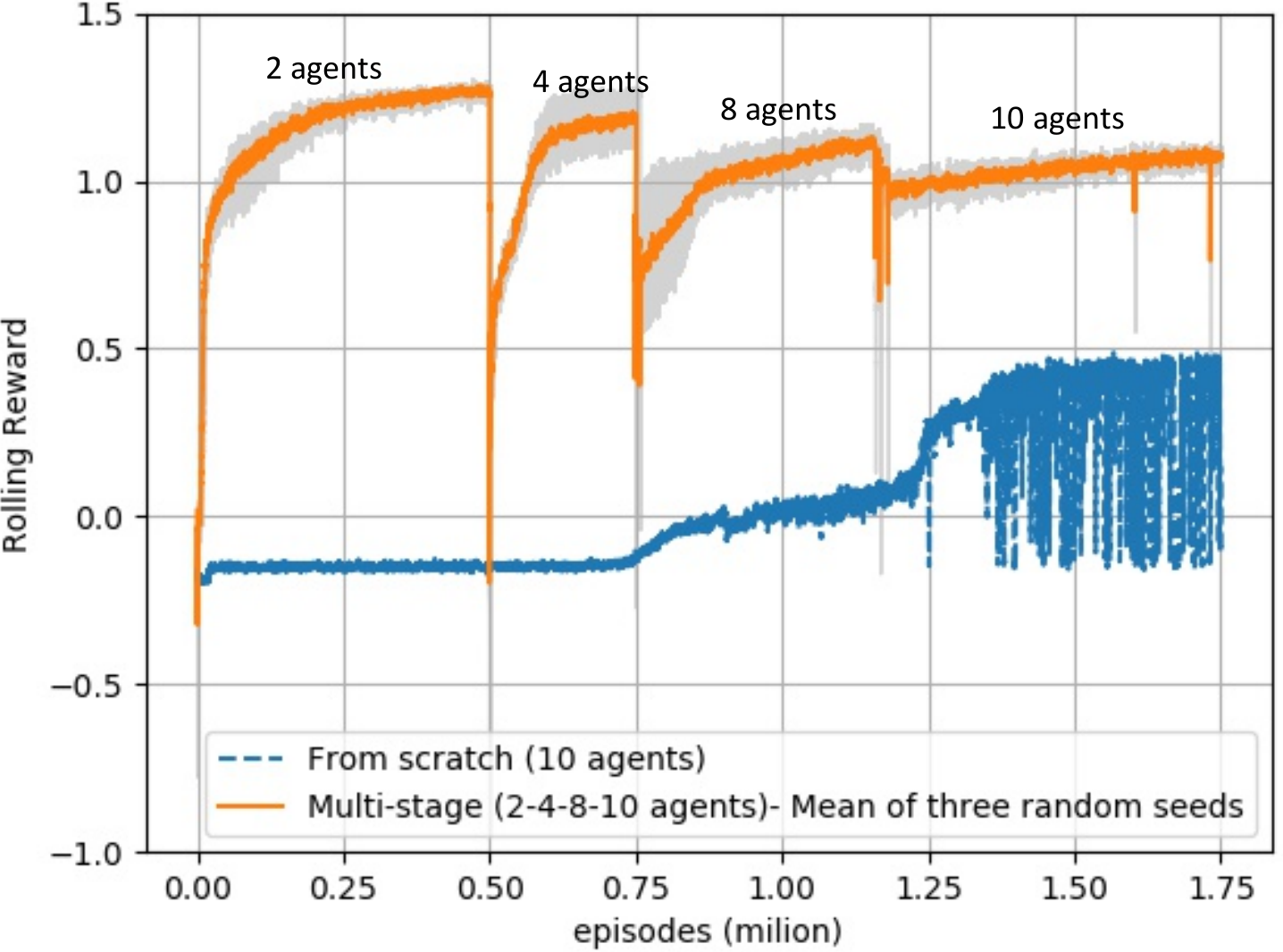}
\caption{Multi-stage training used for training GA3C-CADRL-NSL algorithm and its comparison with training from scratch. Rolling reward is the average reward over last 1000 episodes.}
\label{Fig. 1}
\end{figure}

GA3C-CADRL-NSL generates random scenarios to learn an optimal policy. All scenarios include random agents' initial and goal locations. Each agent's radius and preferred velocity selected randomly from a uniform distribution in the following ranges: $r \in[0.2,0.8] \mathrm{m}$ and $v_{\text {pref}} \in[0.5,2.0] \mathrm{m} / \mathrm{s}$. Many different curriculum training paradigms used in the literature. For example, \cite{everett2018motion} starts training with SL, then runs two RL phases: one with 2-4 agents in the environment and next with 4-10 agents. \cite{long2018towards} uses a two-stage training process, the first stage has 20 agents placed randomly in a simple environment without any obstacle. The second stage continues training with 58 agents placed in a richer and more complex environment including obstacles. In this paper we propose a four-stage training process, which accelerates the policy to converge to a satisfying solution, and gets higher rewards than the policy trained from scratch with the same number of epoch (as shown in Fig. \ref {Fig. 1}). As the focus of this research is teaching the agents to move in dense environments, we keep the size of the environment fixed ($8 \times 8$ in 2D and $8 \times 8 \times 4$ in 3D) and increase the number of agents in several stages to make denser and harder problems to solve. The work begins the training phase with 2 agents (no need for SL) so that the policy learns both collision avoidance and goal reaching ideas at the same time for reasonably simple examples. Upon convergence, the second RL phase begins for 4 agents starting from the saved trained policy from the previous RL stage. This work continues for 8 and 10 agents in the next stages. Fig. \ref {Fig. 1} compares the rolling rewards over episode for multi-stage training and training from scratch and shows how multi-stage training can achieve a better reward in a shorter time.  Multi-stage training is repeated with three random seeds and the mean value with shading +/- sigma is reported in Fig. \ref {Fig. 1}.           
\squeezeup
\subsection{DRL-FMP Hybrid Control Framework} \label{sec:3-3}
Although GA3C-CADRL-NSL provides better performance than GA3C-CADRL, as represented in Section \ref{S-S}, however, it still suffers from many stuck scenarios, especially in denser environments. The new reward function in GA3C-CADRL-NSL results in more conservative behavior among agents by increasing the collision penalty to -1. This behavior converts most of the GA3C-CADRL's collision scenarios to stuck scenarios. On the other hand, it is proven that the FMP algorithm is live-lock free and it has shown good performance in avoiding stuck situations \cite{SamaneFMP}. Therefore, to solve stuck scenarios we propose to switch to the FMP algorithm in the case of stuck scenarios and then switch back to GA3C-CADRL-NSL. 
The idea of switching to FMP is motivated by solving stuck situations but expanded to two other scenarios: Simple and High-risk. 

Simple scenarios are those in which the agent has no neighboring agent. Using GA3C-CADRL and its new version GA3C-CADRL-NSL in simple scenarios, the agent wanders around its initial location instead of directly moving toward its goal. The reason is that in the training phase, there was always more than one agent in the neighborhood of each learning agent, so it never learned how to handle an observation with no neighboring agents. To solve this issue, for simple scenarios we switch to FMP which guides the agent directly toward its goal location by calculating the summation of force vectors (only navigational feedback force, no collision avoidance force).
  
High-risk scenarios are those in which the agent is too close to the surrounding agent or obstacle and the risk of collision using learned policy is too high. This situation happens for example when the agent should avoid a dynamic obstacle/agent whose dynamics are far from what the agent has learned to deal with in the training stage or a static obstacle/agent whose shape/radius is far from the what was seen before in the training stage. On the other hand, FMP is able to avoid surrounding agents/obstacles even it is very close to them. Actually, in FMP, repulsive forces become effective when the agent is in very close proximity of other agents or obstacles. Therefore, to avoid collision in high-risk situations we propose to switch to FMP when the agent is very close to collide.    
\begin{figure}[!t] 
\centering
\includegraphics[width=3in]{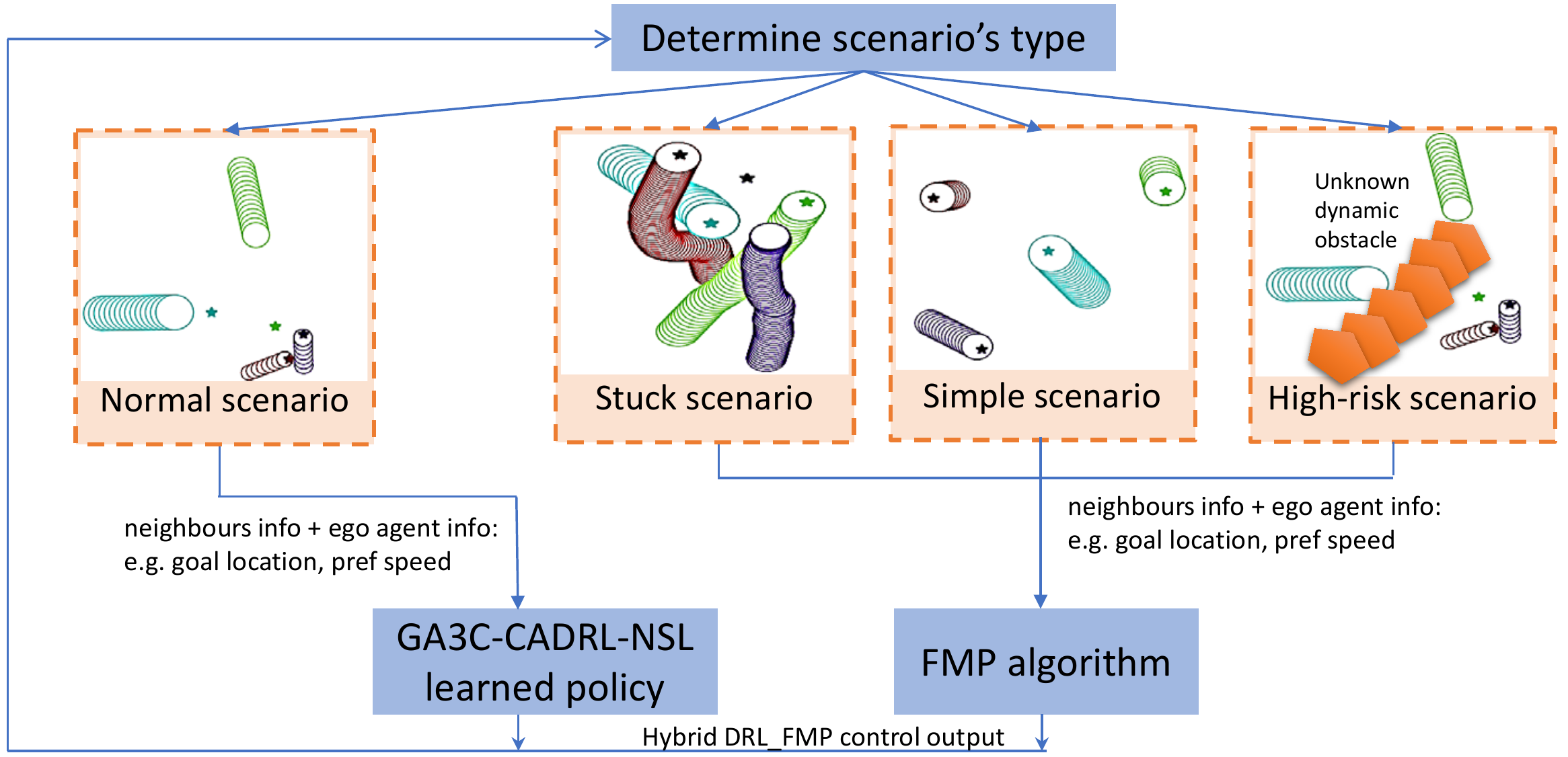}
\caption{DRL-FMP Hybrid Control Framework.}
\label{Fig. 2}
\end{figure}
Figure \ref{Fig. 2} represents the DRL-FMP Hybrid Control Framework described above. As shown in this figure, each agent decides to use FMP or its GA3C-CADRL-NSL learned policy based on its situation (Algorithm \ref{alg}). If it has no neighbours (simple scenario), was too close to others (high-risk scenario) or was in stuck (stuck scenario), it uses FMP to calculate its next action, otherwise (normal scenario) the agent selects its next action using the learned policy in GA3C-CADRL-NSL. The parameters used in Algorithm \ref{alg} are set as follows $r_{FMP}= \sqrt[3]{3 v_{pref}^{2} / 2 \rho}$, $\rho=7.5 \cdot 10^{6}$ and $c_{stuck}=20$. The value of $r_{FMP}$ is selected based on the value is selected in \cite{SamaneFMP} for radius in which FMP algorithm start applying repulsive forces between the agents. This radius is very small and show the case that agents are very close to collide to each other.        $d_{min}$, $|N_i|$ and $\#(time-steps-stay-on-same-place)$ can be calculated by each agent locally as each agent has $\mathbf{s}_{t}^{o}$ information of its neighbors.

\vspace*{-2mm}
\begin{algorithm}
\caption{DRL-FMP Hybrid Controller}\label{alg}
\begin{algorithmic} [1]
\State  \textbf{Input:}  other agent's state, $\mathbf{s}_{t}^{o}$ and goal position, $P_g$
\State  \textbf{Output:} agent’s speed and change in heading angle(s): $u_t$
\If { $d_{min}<r_{FMP}$ \textbf{or} $|N_i|=0$ \textbf{or} \\
 ~ $\#(time-steps-stay-on-same-place) >c_{stuck}$ 
 } \Comment{high-risk, simple, stuck scenarios resp.}\State {$\mathbf{u}_{t} \leftarrow \pi_{FMP}$}
\Else \State {$\mathbf{u}_{t} \leftarrow \pi_{GA3C-CADRL-NSL}$}
\EndIf
\State {return $\mathbf{u}_{t}$ }
\end{algorithmic}
\end{algorithm}  
\vspace*{-7mm}

\section{Simulation Results} \label{S-S}
The GA3C-CADRL-NSL algorithm was implemented in Python using TensorFlow \cite{abadi2016tensorflow}. We executed our implementation on a laptop running Windows 10, with an Intel Core i7-7700HQ 2.8GHz CPU, NVIDIA GeForce GTX 1050 GPU and 16 GB RAM. The step-size $\Delta t$ was set to 0.2$s$ for training and 0.1$s$ for testing just to get more granularity. We found $dt=0.1$ in training made each episode take so long, and very little changes from step to step. The on-line execution of the learned policy runs in real-time and takes on average 0.4-0.5 ms. In total, the RL converges in about 35 hours ($1.8 \cdot 10^{6}$ episodes) for the 2D environment and 53 hours ($5 \cdot 10^{6}$ episodes) for the 3D environment. We compare the results of DRL-FMP Hybrid algorithm with GA3C-CADRL-NSL, GA3C-CADRL \cite{everett2018motion} and FMP algorithms.  All these algorithms are implemented in Python. We also compared the results with Optimal reciprocal collision avoidance (ORCA) \cite{van2011reciprocala} algorithm to show the advantage of proposed method over the established baseline in this field. Our implementation of ORCA builds on open-source ORCA implementation \cite{orca}. To evaluate the performance of the algorithms over dense problems we keep the environment size fixed ($8 \times 8 (m)$ in 2D and $8 \times 8 \times 4(m)$ in 3D) and increase the number of agents to make denser and harder problems. For each density, 50 test cases are generated with random initial and final position, radius, and preferred velocity. All the algorithms are evaluated over the same test cases and the average results are reported. The algorithms are compared over the following parameters: percentage of successful scenarios, percentage of cases with collision, percentage of cases where an agent gets stuck and does not reach its goal and average extra time to reach the goal (beyond a straight path at $v_{pref}$) for successful scenarios, $\overline{t}_{e}$. These parameters provide a measurement of efficiency and optimality.

\begin{figure}[!t]
  \centering
  
  \subfloat[]{\includegraphics[width=.49\linewidth]{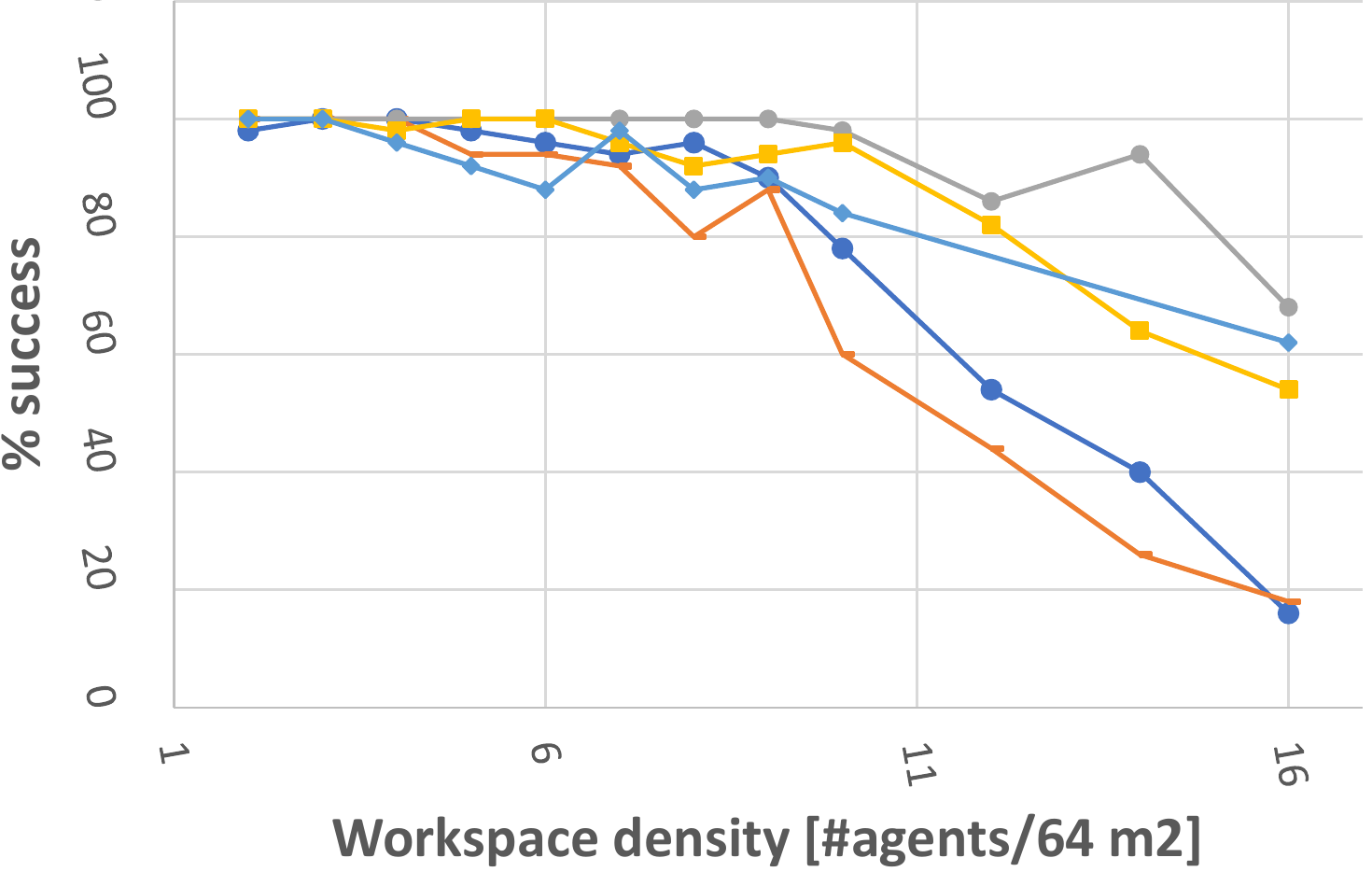}\label{Fig. 3}}
  \hfill
  \subfloat[]{\includegraphics[width=.49\linewidth]{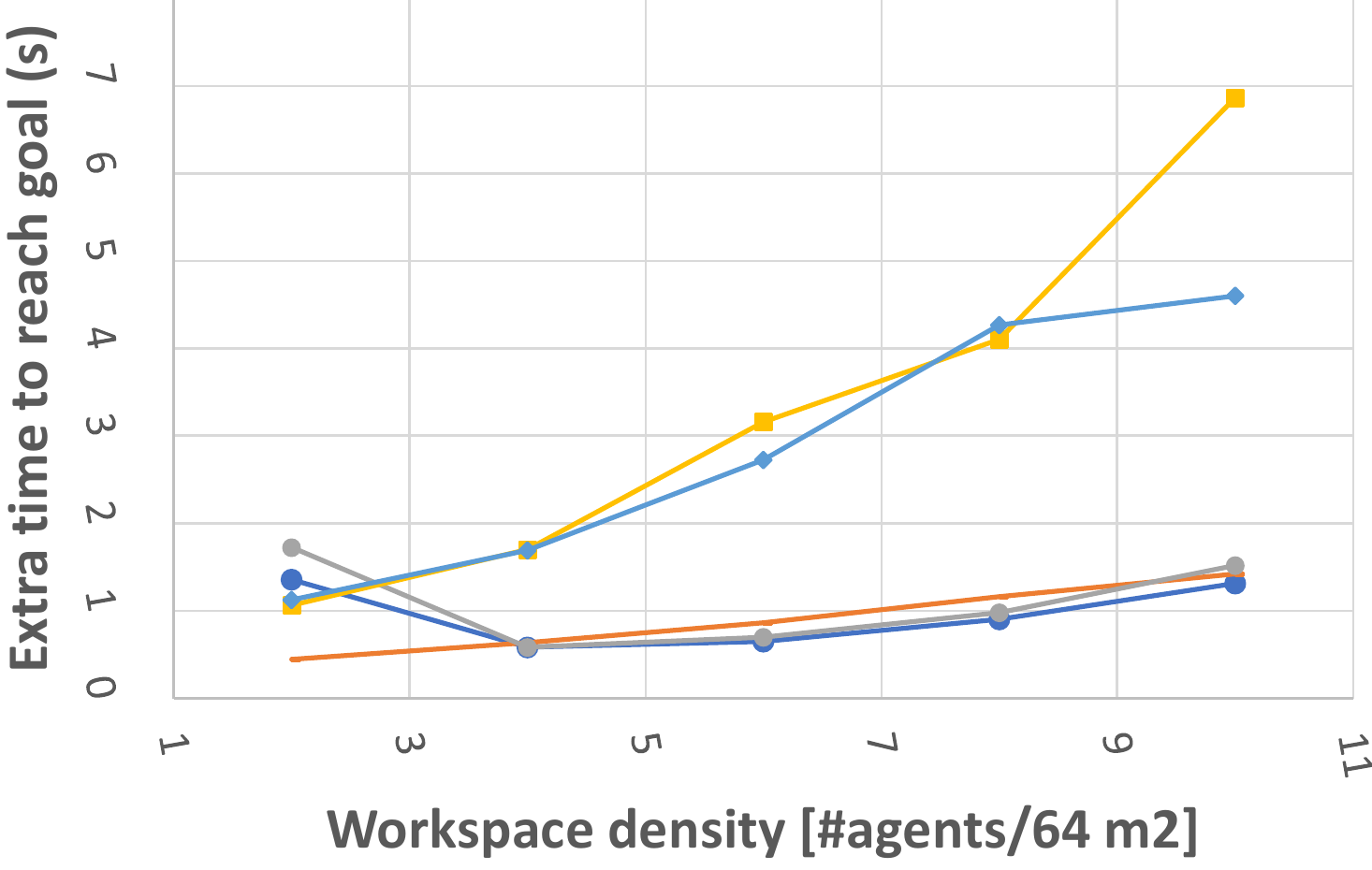}\label{Fig. 4}}
  \hfill
  \subfloat{\includegraphics[width=0.7\linewidth]{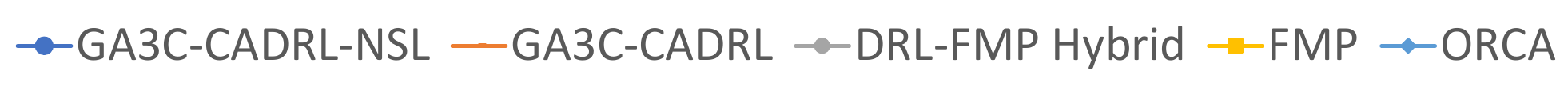}\label{Fig. 4-1}}
  
  \caption{Comparison of percentage of successful scenarios (a), and average extra time to reach goal (b), over different density problems in the 2D environment.}
\end{figure}

Fig \ref{Fig. 3} compares five algorithms over the percentage of successful scenarios on different density problems in the 2D environment. As represented in this figure, all the algorithms work fine for low-density problems, but their performance decreases in denser problems. As expected, DRL-FMP Hybrid algorithm is able to improve the percentage of successful scenarios by up to 50$\%$. Table  \ref{tab. 1} shows the reason for failure (collision/stuck) for each algorithm. as represented in this Figure, the failures in GA3C-CADRL are mostly due to the collision while in GA3C-CADRL-NSL are mostly due to stuck. It is because the reward function in GA3C-CADRL-NSL is more conservative and considers a higher penalty for collision and as a result, most of the collisions are converted to stuck. Then most of the stuck scenarios are solved in DRL-FMP Hybrid algorithm by switching to FMP.        

Fig \ref{Fig. 4} compares five algorithms over average extra time to reach goal over different density problems in the 2D environment. As represented in this Figure FMP has the worst performance and all the others are very close to each other. It is what was expected as FMP is a distributed algorithm in which none of the agents have a general view of the problem and is not able to provide an optimal path.

We expand GA3C-CADRL-NSL and DRL-FMP Hybrid algorithms to the 3D environment. The output command is $u_{t}=\left[v_{t}, \theta_{t}, \varphi_{t}\right]$ (speed, change in heading angle (polar $\theta_{t}$ and azimuthal $\varphi_{t}$)) in this environment which is discretized into 43 actions instead of 11 actions originally defined for 2D in GA3C-CADRL. The environment size is fixed to $8 \times 8 \times 4 (m)$ and the number of agents increased from 2 to 16. GA3C-CADRL-NSL trained using multi-stage training with 2-4-8-10-16 agents and the results converged after 5 million episodes. After stage 1 (converges in 1 $\cdot 10^{6}$ episodes), they average 1.35 reward per episode. When RL phase 2 begins, the rolling reward drops to 0.7 initially since the domain becomes much harder, and then increases until converging at 1.3 (after a total of 1.5 $\cdot 10^{6}$ episodes). Similar behavior repeats for 8, 10 and 16 agents. Finally, 16 agents converge to an average of 1.17 per episode. We experience higher rewards in 3D in comparison to the same number of agents in 2D because the environment is 4 times larger and as a result, the density is lower. Evaluation results confirm the results reported for the 2D environment and are represented in Table \ref{tab. 2}. DRL-FMP Hybrid algorithm provides the highest percentage of successful scenarios and the lowest average extra time to reach the goal.  
\begin{table}[]
\scriptsize
\caption {Comparison of percentage of failures (collisions or stuck) over different density problems in 2D environment} \label{tab. 1} 
\setlength{\tabcolsep}{3pt}
\begin{tabular}{|c|c|c|c|c|c|}
\hline
\multicolumn{2}{|c|}{Test case setup} & \multicolumn{4}{c|}{$\%$ failure ($\%$ collisions / $\%$ stuck)} \\ \hline
\#ags. & Size(m) & DRL-FMP Hyb. & GA3C-CADRL-NSL & GA3C-CADRL & FMP \\ \hline
2 & $8 \times 8 $ & 0 (0/ 0) & 2 (0/ 2) & 0 (0/ 0) & 0 (0/ 0) \\ \hline
3 & $8 \times 8 $ & 0 (0/ 0) & 0 (0/ 0) & 0 (0/ 0) & 0 (0/ 0) \\ \hline
4 & $8 \times 8 $ & 0 (0/ 0) & 0 (0/ 0) & 0 (0/ 0) & 2 (0/ 2) \\ \hline
5 & $8 \times 8 $ & 0 (0/ 0) & 2 (0/ 2) & 6 (4/ 2) & 0 (0/ 0) \\ \hline
6 & $8 \times 8 $ & 0 (0/ 0) & 4 (0/ 4) & 6 (4/ 2) & 0 (0/ 0) \\ \hline
7 & $8 \times 8 $ & 0 (0/ 0) & 6 (0/ 6) & 8 (6/ 2) & 4 (2/ 2) \\ \hline
8 & $8 \times 8 $ & 0 (0/ 0) & 4 (0/ 4) & 20 (20/0) & 8 (6/ 2) \\ \hline
9 & $8 \times 8 $ & 2 (2/ 0) & 10 (0/10) & 12 (12/0) & 6 (4/ 2) \\ \hline
10 & $8 \times 8 $ & 2 (2/ 0) & 22 (0/22) & 40 (30/10) & 4 (4/ 0) \\ \hline
12 & $8 \times 8 $ & 12 (8/ 4) & 46 (2/44) & 56 (46/10) & 18 (12/6) \\ \hline
14 & $8 \times 8 $ & 28 (22/6) & 60 (2/58) & 74 (56/18) & 36 (22/14) \\ \hline
16 & $8 \times 8 $ & 38 (24/14) & 84 (4/80) & 82 (74/8) & 46 (18/28) \\ \hline
\end{tabular}
\vspace*{+3mm}
\end{table}      

\begin{table}[]
\scriptsize
\caption {Comparison of $\%$ successful scenarios and avg. extra time to goal over different density problems in 3D} \label{tab. 2} 
\setlength{\tabcolsep}{2pt}
\begin{tabular}{|c|c|c|c|c|c|c|c|}
\hline
\multicolumn{2}{|c|}{Test case setup} & \multicolumn{2}{c|}{DRL-FMP Hyb.} & \multicolumn{2}{c|}{GA3C-CADRL-NSL}& \multicolumn{2}{c|}{FMP} \\ \hline
\#ags. & Size(m) & $\%$ success & Avg $\overline{t}_{e}$(s) & $\%$ success & Avg $\overline{t}_{e}$(s) & $\%$ success & Avg $\overline{t}_{e}$(s) \\ \hline
2 & $8 \times 8  \times 4 $ & 100 & 1.77 & 100 & 1.77 & 100 & 0.85 \\ \hline
3 & $8 \times 8  \times 4 $ & 100 & 1.07 & 100 & 1.18 & 100 & 0.92 \\ \hline
4 & $8 \times 8  \times 4 $ & 100 & 0.56 & 100 & 0.52 & 100 & 1.04 \\ \hline
5 & $8 \times 8  \times 4 $ & 100 & 0.45 & 100 & 0.44 & 100 & 0.96 \\ \hline
6 & $8 \times 8  \times 4 $ & 100 & 0.46 & 100 & 0.47 & 100 & 1.19 \\ \hline
7 & $8 \times 8  \times 4 $ & 100 & 0.56 & 100 & 0.61 & 100 & 1.42 \\ \hline
8 & $8 \times 8  \times 4 $ & 100 & 0.51 & 100 & 0.53 & 98 & 1.47 \\ \hline
9 & $8 \times 8  \times 4 $ & 100 & 0.54 & 98 & 0.60 & 100 & 1.60 \\ \hline
10 & $8 \times 8  \times 4 $ & 100 & 0.64 & 98 & 0.70 & 98 & 1.77 \\ \hline
16 & $8 \times 8  \times 4 $ & 100 & 0.77 & 84 & 0.89 & 98 & 2.38 \\ \hline
\end{tabular}
\vspace*{-7mm}
\end{table}
\squeezeup
\section{Conclusions} 
This work presented GA3C-CADRL-NSL and its combination with FMP that is called DRL-FMP Hybrid algorithm to solve a distributed motion planning problem in dynamic, dense and crowded environments. GA3C-CADRL-NSL improves the performance of recent GA3C-CADRL algorithm by introducing a new reward function that not only eliminates the requirement of a pre supervised learning step but also decreases the chance of collision in crowded environments. The new reward function is more conservative and as a result, converts many collision scenarios to the stuck situation. Then DRL-FMP Hybrid algorithm solves most of the stuck scenarios by switching to FMP. Switching to FMP is not limited to the stuck situation. It also includes high-risk and simple scenarios. Both 2D and 3D simulation results show DRL-FMP Hybrid algorithm outperforms both GA3C-CADRL and FMP algorithms in terms of percentage of successful scenarios and time-optimality of the produced path. For future work, we are looking to implement this algorithm on real robots. Also, we are looking to expand the proposed algorithm to generate continuous output instead of current discrete ones to increase the flexibility of agents movements. Expanding this algorithm to take into account agents with more complicated kinematic constraints can be considered as another future work for this research.     
   
\squeezeup
\squeezeup
\bibliographystyle{IEEEtran}
\bibliography{ref}

%

\end{document}